\documentclass[openreview,10pt]{IEEEtran}
\IEEEoverridecommandlockouts
\usepackage{cite}
\usepackage{amsmath,amssymb,amsfonts}
\usepackage{algorithmic}
\usepackage{graphicx}
\usepackage[lofdepth,lotdepth]{subfig}
\usepackage[font=footnotesize,labelfont=bf]{caption}
\usepackage{textcomp}
\usepackage{xcolor}
\usepackage{float}
\usepackage{enumitem}
\usepackage{booktabs}
\usepackage{multirow}
\usepackage[square,numbers]{natbib}
\usepackage{cleveref}

\def\BibTeX{{\rm B\kern-.05em{\sc i\kern-.025em b}\kern-.08em
    T\kern-.1667em\lower.7ex\hbox{E}\kern-.125emX}}

\usepackage{fancyhdr,lipsum}
\fancypagestyle{firstpage}{% Page style for first page
  \fancyhf{}% Clear header/footer
  \fancyhead[C]{To appear at the 23rd Design, Automation and Test in Europe (DATE 2020). Grenoble, France.}% Header
  \fancyfoot[C]{\thepage}% Footer
}

\begin{document}

\title{ ReD-CaNe: A Systematic Methodology for
Resilience Analysis and Design of Capsule
Networks under Approximations
%\vspace{-0.7em}
}
%\vspace*{-17.8mm}}

\author{\IEEEauthorblockN{Alberto Marchisio\IEEEauthorrefmark{1}, Vojtech Mrazek\IEEEauthorrefmark{2}, Muhammad Abudllah Hanif\IEEEauthorrefmark{1}, Muhammad Shafique\IEEEauthorrefmark{1}\\}
\begin{small}
\IEEEauthorblockA{\IEEEauthorrefmark{1}Faculty of Informatics, Embedded Computing Systems, Technische Universit\"at Wien (TU Wien), Vienna, Austria \\
\IEEEauthorrefmark{2}Faculty of Information Technology, IT4Innovations Centre of Excellence, Brno University of Technology, Czech Republic\\
Email: \{alberto.marchisio, muhammad.hanif, muhammad.shafique\}\@tuwien.ac.at, mrazek@fit.vutbr.cz
}
\end{small}
\vspace*{-5mm}
}

\maketitle
\thispagestyle{firstpage}

\begin{small}
\begin{abstract}
Recent advances in Capsule Networks (CapsNets) have shown their superior learning capability, compared to the traditional Convolutional Neural Networks (CNNs). However, the extremely high complexity of CapsNets limits their fast deployment in real-world applications. Moreover, while the resilience of CNNs have been extensively investigated to enable their energy-efficient implementations, the analysis of CapsNets' resilience is a largely unexplored area, that can provide a strong foundation to investigate techniques to overcome the CapsNets' complexity challenge.

Following the trend of Approximate Computing to enable energy-efficient designs, we perform an extensive resilience analysis of the CapsNets inference subjected to the approximation errors. Our methodology models the errors arising from the approximate components (like multipliers), and analyze their impact on the classification accuracy of CapsNets. This enables the selection of approximate components based on the resilience of each operation of the CapsNet inference. We modify the TensorFlow framework to simulate the injection of approximation noise (based on the models of the approximate components) at different computational operations of the CapsNet inference. Our results show that the CapsNets are more resilient to the errors injected in the computations that occur during the dynamic routing (the softmax and the update of the coefficients), rather than other stages like convolutions and activation functions. Our analysis is extremely useful towards designing efficient CapsNet hardware accelerators with approximate components. To the best of our knowledge, this is the first proof-of-concept for employing approximations on the specialized CapsNet hardware.

%Other than inserting random noise, we model the error patterns based on use case scenarios, like using approximate hardware, applying quantization and memory related errors. Our analysis shows that TODO and TODO has more impact on the test accuracy, while TODO is more resilient thanks to TODO. To the best of our knowledge, it is the first proof of concept for employing approximations on CapsNet hardware.
\end{abstract}


\vspace*{-2mm}
\end{small}

\section{Introduction}
\vspace*{-1mm}

The recently introduced breakthrough concept of Capsule Networks (CapsNets) by the Google Brain team has achieved a significant spotlight due to its powerful new features offering high accuracy and better learning capabilities \cite{Sabour2017dynamic_routing}. Traditional Convolutional Neural Networks (CNNs) are not able to learn the spatial relations in the images much efficiently \cite{Sabour2017dynamic_routing}. Moreover, they make extensive use of the pooling layer to reduce the dimensionality of the space, and consequently as a drawback, the learning capabilities are reduced. Therefore, a huge amount of training data is required to mitigate such deficit. On the other hand, \textit{CapsNets take advantage from their novel structure, with so-called capsules and their cross-coupling learnt through the dynamic routing algorithm, to overcome this problem}. Capsules produce vector outputs, as opposed to scalar outputs of CNNs \cite{Sabour2017dynamic_routing}. In a vector format, CapsNets are able to learn the spatial relationships between the features. For example, the Google Brain team \cite{Sabour2017dynamic_routing} demonstrated that CNNs recognize an image where the nose is below the mouth as a ``face'', while CapsNets do not make such mistake because they have learned the spatial correlations between features (e.g., the nose must appear above the mouth). Other than image classification, CapsNets have been successfully showcased to perform vehicle detection \cite{Yu2019CapsNetDetection}, speech recognition \cite{Wu2010CapsNetspeechrecognition} and natural language processing \cite{zhao2019CapsNetNLP}.

The biggest roadblock in the real-world deployments of CapsNet inference is their extremely high complexity, requiring a specialized hardware architecture (like the recent one in \cite{Marchisio2019CapsAcc}) that may consume a significant amount of energy/power. Not only deep CapsNet models \cite{Rajasegaran2019DeepCaps}, but also shallow models like \cite{Sabour2017dynamic_routing} require intense computations due to matrix multiplications in the capsule processing and the iterative dynamic routing algorithm for learning the cross-coupling between capsules. To deploy CapsNets at the edge, as commonly adopted for the traditional CNNs \cite{Han2016EIE}, network compression techniques (like pruning and quantization) \cite{Han2016DeepCompression} can be applied, but at the cost of some accuracy loss. Moreover, the current trends in Approximate Computing can be leveraged to achieve energy-efficient hardware architectures, as well as for enabling design-/run-time energy-quality tradeoffs. However, this requires a comprehensive resilience analysis of CapsNets considering approximation errors in the hardware, in order to make correct design decisions on which computational steps of the CapsNets are more likely to be approximated and which not. Note, unlike in the case of approximations, an error can also be caused by a misfunctioning of the computing hardware \cite{Jiao2017VulnerabilityDNN} or of the memory \cite{Kim2014RowHammer}. Fault injections have demonstrated to fool CNNs \cite{Liu2017FaultInjectionDNN}, and can potentially cause a CapsNet misclassification as well.

\textbf{Concept Overview and our Novel Contributions:}

\begin{figure}[b]
	\centering
	\vspace*{-7mm}
	\includegraphics[width=.95\linewidth]{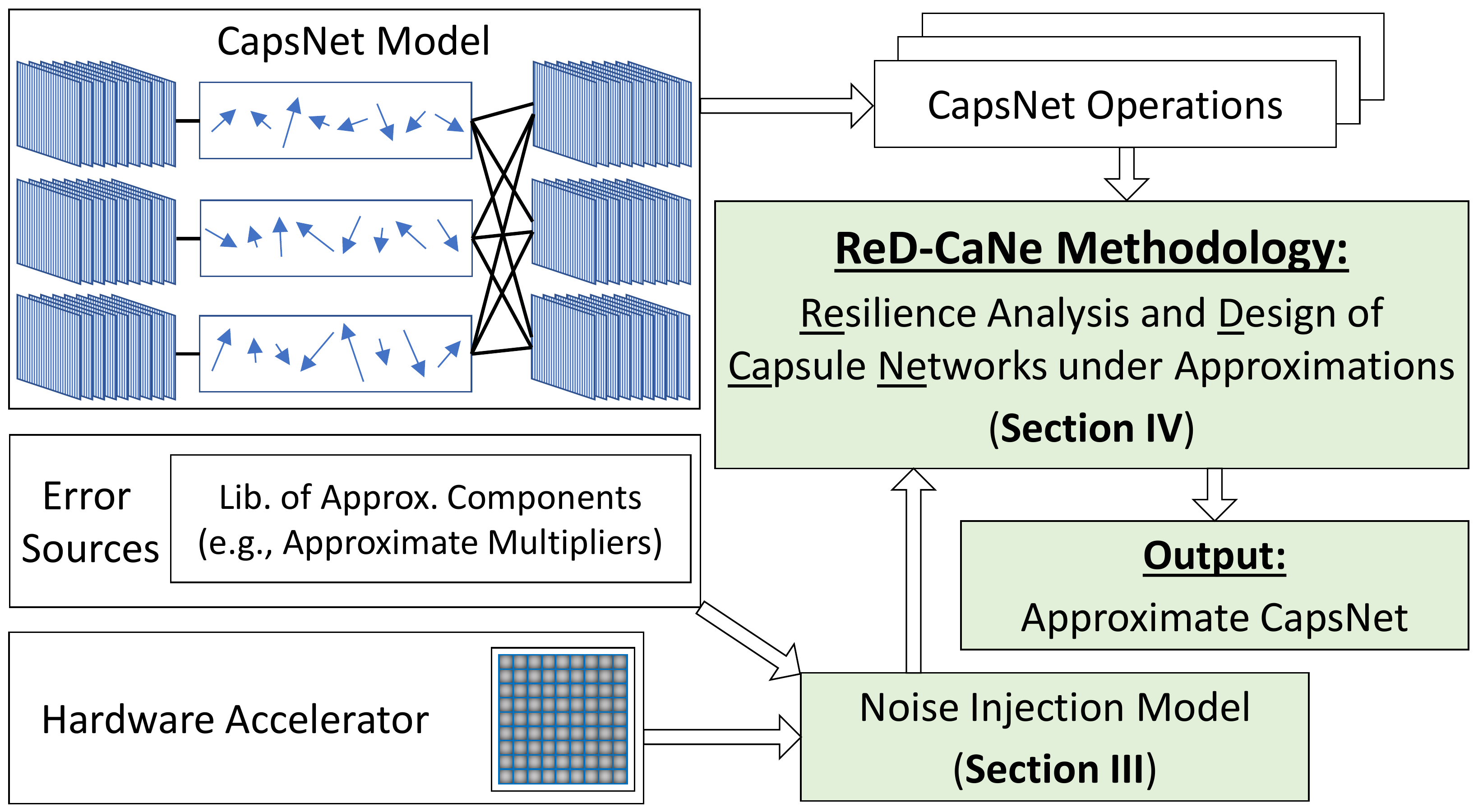}
	\vspace*{-2mm}
	\caption{Overview of our ReD-CaNe methodology. The novel contributions are shown in green boxes.}
	\label{fig:overview}
	\vspace*{-2mm}
\end{figure}

\begin{figure*}[t]
	\centering
	\vspace*{-0mm}
	\includegraphics[width=.9\linewidth]{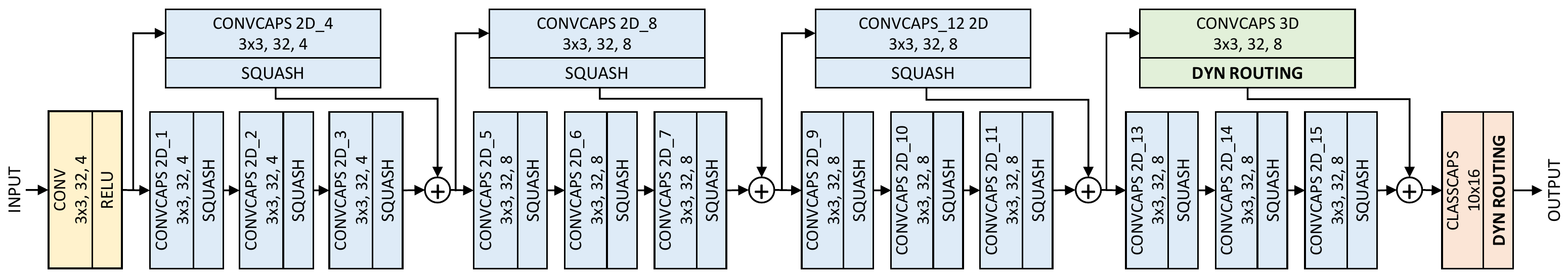}
	\vspace*{-2mm}
	\caption{The architecture of the DeepCaps network \cite{Rajasegaran2019DeepCaps}.}
	\label{fig:deepcaps}
	\vspace*{-3mm}
\end{figure*}

To address these challenges, we propose \textit{ReD-CaNe}, a novel methodology (see Fig. \ref{fig:overview}) for analyzing the resilience of CapsNets under approximations, which, to the best of our knowledge, is \textit{the first of its kind}. First, we devise a noise injection model to simulate real-case scenarios of errors coming from approximate hardware components like multipliers, which are very common in multiply-and-accumulate (MAC) operations for the matrix multiplications of capsules. Then, we analyze the error resilience of the CapsNets by building a systematic methodology for injecting noise into different operations of the CapsNet inference, and evaluating their impact on the accuracy. The outcome of such analysis will produce guidelines for designing and selecting approximate components, based on the resilience of each operation. At the output, our methodology produces an approximated version of a given CapsNet, to achieve an energy-efficient inference.

\vspace*{-1mm}
\subsection{In a nutshell, our novel contributions are:}

\begin{itemize}[leftmargin=*]
    \item We analyze and \textit{model the noise injections} that can be generated by different approximate arithmetic components, e.g., multipliers. (\textbf{Section \ref{sec:model_errors}}) % \vm{or adders.} \vmR{Vojta, if we are just approximating the multipliers, why should we write adders here? It can confuse the reviewers}
    \item We devise \textit{ReD-CaNe}, a novel methodology for analyzing the \underline{Re}silience and \underline{D}esigning \underline{Ca}psule \underline{Ne}tworks under approximations, by systematically adding noise at different operations of the CapsNet inference and by monitoring the test accuracy. The approximated components are selected based on the resilience level of the different operations of the CapsNet inference. (\textbf{Section \ref{sec:ERCaNe_methodology}})
    \item We test our methodology on several benchmarks. On the DeepCaps model \cite{Rajasegaran2019DeepCaps} for CIFAR-10 \cite{Krizhevsky2009CIFAR}, MNIST \cite{LeCun1998MNIST}, and SVHN \cite{Netzer2011SVHN} datasets, and on the CapsNet model \cite{Sabour2017dynamic_routing} for MNIST and Fashion-MNIST \cite{Xiao2017Fashion-MNIST} datasets. Our results demonstrate that the least resilient operations are the convolutions in CapsLayers, while \textit{the operations performed during the dynamic routing of the Caps3D and ClassCaps layers are relatively more resilient}. (\textbf{Section \ref{sec:results}})
\end{itemize}

Before proceeding to the technical sections, in \textbf{Section \ref{sec:background}}, we summarize the concepts of CapsNets and review the existing works of error resilience for traditional CNNs, with necessary details to understand the rest of the paper.

%There can be many sources of error, unintentional (faults in the hardware or in the memory) or intentional (approximations, pruning, quantization, weight sharing).

%Each of them has a different pattern.

%Since there is the routing-by-agreement, this can have an influence on the error resilience of CapsNets.

%Existing works analyzed the error resilience of traditional DNNs.

%We develop a methodolody to systematically analyze the error resilience of the CapsNets, in order to bear extensive use of approximations with minimal loss of accuracy. And this is also useful for the hardware fault tolerance.

%We test our methodology on Sabour et al. \cite{Sabour2017dynamic_routing}, GTSRB and DeepCaps.

\vspace*{-1mm}
\section{Background and Related Work}
\label{sec:background}
\vspace*{-1mm}

\subsection{Capsule Networks (CapsNets)}
\label{subsec:background_capsnets}

CapsNets, first introduced in \cite{Hinton2011TransformingAutoencoder}, have become popular in \cite{Sabour2017dynamic_routing}, thanks to the new concepts like capsules and the dynamic routing algorithm. Following this trend, DeepCaps \cite{Rajasegaran2019DeepCaps} proposed to increase the depth of CapsNets, achieving state-of-the-art accuracy for the CIFAR10 \cite{Krizhevsky2009CIFAR} dataset.

A \textit{capsule} is a group of neurons where the instantiation parameters are represented as the orientation of each element of the vector, and the vector length represents the probability that the entity exists. Moreover, vector predictions of the capsules need to be supported by nonlinear vectorized activation functions. Towards this end, the squashing function bounds the output of the capsule between 0 and 1.

\begin{figure}[h]
	\centering
	\vspace*{-1mm}
	\includegraphics[width=.95\linewidth]{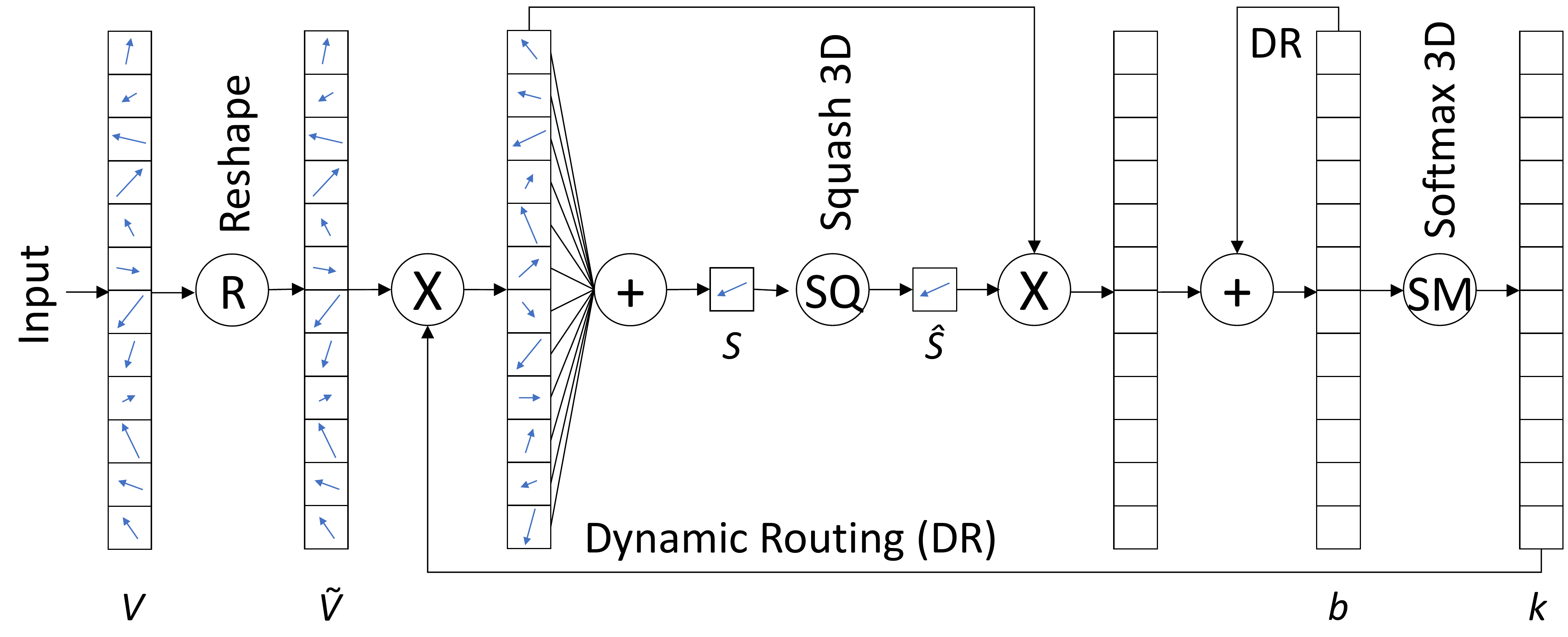}
	\vspace*{-2mm}
	\caption{The operations to be computed for the dynamic routing.}
	\label{fig:routing_by_agreement}
	\vspace*{-3mm}
\end{figure}

In the dynamic routing, \textit{the coupling coefficients}, which are connecting two consecutive capsule layers, \textit{learn the agreement during the inference} by iteratively updating their values according to the relevance of the path. As an example, the architecture\footnote{Since we focus on the CapsNet inference, we do not discuss the operations that are involved in the training process {\em only} (e.g., decoder and reconstruction loss). For further details on CapsNets, we refer the readers to \cite{Sabour2017dynamic_routing}\cite{Rajasegaran2019DeepCaps}.} of the DeepCaps is shown in the Fig. \ref{fig:deepcaps}. It has 16 convolutional capsule layers (ConvCaps), where one of them is 3D, and one fully-connected capsule layer (ClassCaps) at the end. A special focus is on the operations required for the dynamic routing, which is performed in the 3D ConvCaps and in the ClassCaps layers, as shown in Fig. \ref{fig:routing_by_agreement}. Note that the operations like matrix-vector multiplications and squash are different from the traditional CNNs. Hence, \textit{a challenge that we want to address in this paper is to study the inter-relation between the precision of these operations and the accuracy of the CapsNets, when subjected to errors due to approximations}.

\subsection{Error Resilience of Traditional CNNs}
\label{subsec:background_resilienceCNN}

\vspace*{-1mm}

The resilience of traditional CNNs has recently been investigated in the literature. \citeauthor{Du2014ErrorResilientAccelerators}~\cite{Du2014ErrorResilientAccelerators} analyzed the error resilience, showing that it is possible to obtain high energy savings with minimal accuracy loss. \citeauthor{Hanif2018ErrorResilienceCNN}~\cite{Hanif2018ErrorResilienceCNN} proposed a methodology to apply approximations for CNNs, based on the error resilience. \citeauthor{Li2017ErrorPropagationDNN}~\cite{Li2017ErrorPropagationDNN} studied the error propagation with the end goal of adopting countermeasures for obtaining resilient hardware accelerators. \citeauthor{Zhang2019faulttolerantDNN}~\cite{Zhang2019faulttolerantDNN} proposed a method to design fault tolerant systolic array-based accelerators. \citeauthor{Hanif2019CANN}~\cite{Hanif2019CANN} introduced a method for applying approximate multipliers into CNN accelerators without any error in the final result. \citeauthor{Mrazek2019autoAx}~\cite{Mrazek2019autoAx}\cite{Mrazek2019ALWANN} proposed a methodology to successfully search and select approximate components for CNN accelerators. \citeauthor{Hanif2018X-DNNs}~\cite{Hanif2018X-DNNs} and \citeauthor{Marchisio2019DL4EC}~\cite{Marchisio2019DL4EC} analyzed cross-layer approximations for CNNs. However, these works analyzed only traditional CNN accelerators, and such studies cannot be efficiently extrapolated for CapsNets, as discussed before. \textit{Hence, there is a dire need to perform the resilience analysis for CapsNets in a systematic way, such that we can take efficient decisions about approximating the appropriate operations of CapsNets.}

%Hence, \textit{the resilience of capsule layers has not been explored yet}. Therefore, we aim at obtaining dedicated approximations for different operations of the CapsNets, which offers more \vm{flexibility} then CNNs due to the specialized computations occurring in the dynamic routing.

\vspace*{-1mm}

\subsection{Error Sources}

\vspace*{-1mm}

In a generic Deep Learning application, errors may occur due to different sources like software approximations (e.g., quantization), hardware approximations (e.g., approximate multipliers), transient faults (i.e., bit flips due to particle strikes) and permanent faults (e.g., stuck-at-zero and stuck-at-one). In this paper, due to the focus on energy-efficiency, we target approximation errors\footnote{For further details on reliability and security related works on DNNs that study soft errors, permanent faults, and adversarial noise, we refer the readers to \cite{Goodfellow2015explainingadvexamples}\cite{Jiao2017VulnerabilityDNN}\cite{Srinivasan20166TSRAMANN}\cite{Zhang2019RobustML}.}.
%different sources of errors might occur. One category is composed of soft errors, which can be generated adversarially \cite{Goodfellow2015explainingadvexamples} or by the presence of noise in the environment (input or output of the network). Another class of errors can be identified by errors due to the underlying computing hardware \cite{Jiao2017VulnerabilityDNN}, e.g., stuck-at faults due to fabrication errors, aging, voltage underscaling \cite{Srinivasan20166TSRAMANN}, or memory-related errors when storing/reading values to/from the memory \cite{Kim2014RowHammer}. %Errors due the usage of approximate components~\cite{Mrazek2017Lib} show their own patterns.

If the CapsNet inference is performed by specialized hardware accelerators \cite{Marchisio2019CapsAcc}, a fixed-point representation is typically preferred, as compared to the floating-point counterpart \cite{google2017quant}. Therefore, a floating-point value $x$, which must be represented in a $b$-bit fixed-point arithmetic~\cite{Parashar2010WLOptimization}, is mapped to a range $[0:s^b-1]$. The quantization function $Q$ is defined in Eq. \ref{eq:quant}.

\vspace*{-0mm}
\begin{equation}
    Q(x)=\frac{x-\mathrm{min}(x) }{\max(x) - \mathrm{min}(x)} \cdot (2^b-1)
    \label{eq:quant}
\end{equation}
\vspace*{-0mm}

%\vm{In the accelerators of inference path of neural networks, fixed-point representation is typically used instead of the floating-point in the training~\cite{google2017quant}. For $b$-bit arithmetic, the float values $X$ must be mapped to a range $0 \dots 2^b-1$. Then the quantization function $Q$ is defined as follow
%$$Q(X)=\frac{X-\mathrm{min}(X) }{\max(X) - \mathrm{min}(x)} \cdot (2^b-1).$$}
In this work, we simulate the CapsNets with floating-point arithmetic, but \textit{the behavior of approximate fixed-point components is simulated by adjusting their values according to the quantization effect}. Hence, we focus on modeling the errors subjected to the employment of approximate components in CapsNet hardware accelerators.

\vspace*{-1mm}

\section{Modeling the Errors as Injected Noise}
\label{sec:model_errors}

\vspace*{-1mm}

\subsection{Analysis of Different Operations in CapsNets}
\label{subsec:motivation_multiplier}

\vspace*{-1mm}

\begin{figure}[h]
\centering
\vspace*{-3mm}
\begin{minipage}[t]{.64\linewidth}
\begin{table}[H]
\centering
\caption{Number and unit energy consumption of different basic operation of the DeepCaps \cite{Rajasegaran2019DeepCaps}.}
\label{tab:energy_operator}
\resizebox{\textwidth}{!}{%
\begin{tabular}{|c|c|c|}
\hline
\textbf{OPERATION} & \textbf{\# OPS} & \multicolumn{1}{l|}{\textbf{Unit Energy {[}pJ{]}}} \\ \hline
Addition & 1.91 G & 0.0202 \\ \hline
Multiplication & 2.15 G & 0.5354 \\ \hline
Division & 4.17 M & 1.0717 \\ \hline
Exponential & 175 K & 0.1578 \\ \hline
Square Root & 502 K & 0.7805 \\ \hline
\end{tabular}%
}
\end{table}
%\vspace*{-2mm}
\end{minipage}
\hfill
\begin{minipage}[t]{.32\linewidth}
\begin{figure}[H]
\centering
\vspace*{-3mm}
\includegraphics[width=.6\linewidth]{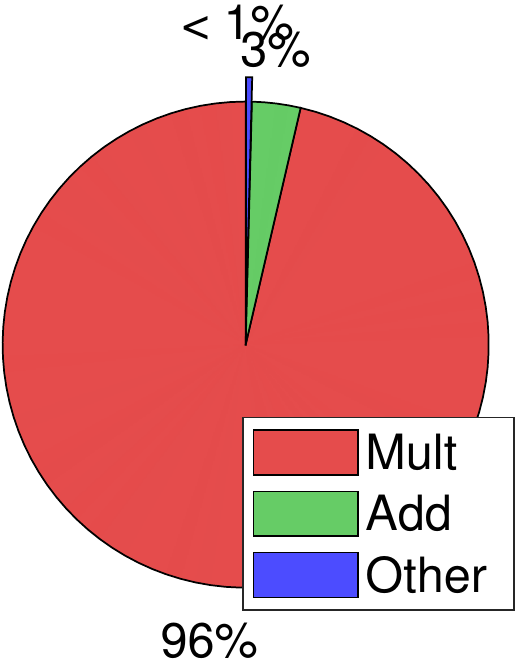}
\caption{Energy breakdown for different ops.}
\label{fig:op_breakdown}
\end{figure}
\end{minipage}
\vspace*{-1mm}
\end{figure}

We perform a comprehensive analysis to investigate which hardware components have the highest impact on the total energy consumption of the CapsNets' computational blocks. Table \ref{tab:energy_operator} reports the number of operations that occur in the computational path of the DeepCaps \cite{Rajasegaran2019DeepCaps} inference and the energy consumption per operation. The latter has been generated by synthesizing the implementation with 8 bits fixed-point operations, in a 45nm CMOS technology with the Synopsys Design Compiler tool. Fig. \ref{fig:op_breakdown} presents the breakdown of the estimated energy share for each operation. It is worth noticing that the multipliers count for 96\% of the total energy share of the computational path of the DeepCaps. The occurrences of the addition is also high, but energy-wise the additions consume only 3\% of the total share due to their reduced complexity as compared to that of the multipliers. Hence, it is important to explore the energy savings from approximating the multiplier operations first, as we target in this paper.

\begin{figure}[h]
\centering
\vspace*{-2mm}
\includegraphics[width=.8\linewidth]{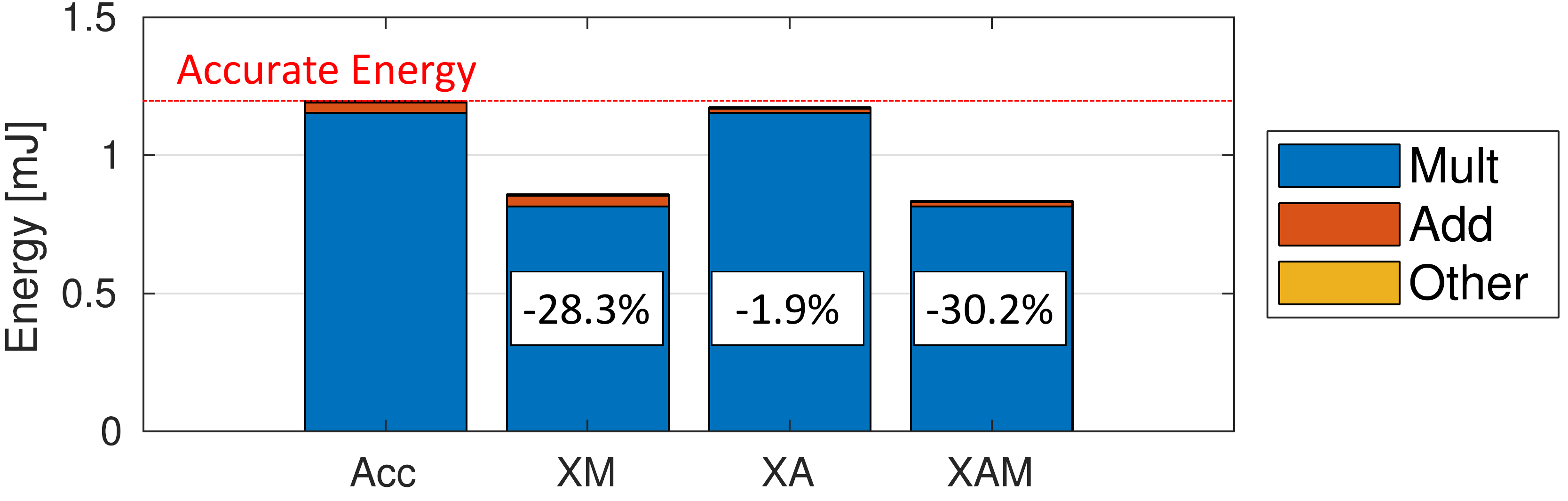}
\vspace*{-2mm}
\caption{Optimization potential by applying approx. components in CapsNets.}
\label{fig:optim_potential}
\vspace*{-2mm}
\end{figure}

In the following, we study the energy optimization potential of employing approximate components. As a case study, we select from the EvoApprox8B library~\cite{Mrazek2017Lib} the NGR approximate multiplier and the 5LT approximate adder. The results in Fig. \ref{fig:optim_potential} show that approximating only the multipliers (XM) can save more than 28\% of the energy consumption, compared to the accurate implementation (Acc). Due to the low share of energy consumed by the additions, the advantage of employing approximate adders (XA) or employing approximate adders and multipliers (XAM) is negligible compared to Acc and XM solutions, respectively.

Motivated by the above discussions and analysis, in the following, without loss of generality and for the ease of proof-of-concept development, \textit{we focus our analyses on the approximate multipliers, since they have high impact on the energy consumption, thus opening huge optimization potentials}.

\vspace*{-1mm}
\subsection{Error Profiles for the Approximate Hardware Multipliers}
\vspace*{-.5mm}

%Motivated by the above analysis, in this work we focus on approximate multipliers, because multiplications are widely employed in the computational path, and they count for about \vmR{40 -- 50\%} \vm{to be changed with the results} of the total computational power of hardware accelerators.
We selected 35 approximate multipliers from the EvoApprox8B library~\cite{Mrazek2017Lib} and analyzed the distributions of the erroneous products $P'$ generated by such multipliers, compared to the accurate product $P$ of an 8-bit multiplier (i.e., 16-bit output). The arithmetic error is computed in Eq. \ref{eq:arithmetic_error}, where $a,b$ denotes the inputs to the multipliers from a representative set of inputs $I$.

\vspace*{-0mm}
\begin{equation}
    \Delta_{P'} = \{\forall a,b \in I: P'(a,b) - P(a,b)\}
    \label{eq:arithmetic_error}
\end{equation}
\vspace*{-0mm}

The distributions of the arithmetic errors are calculated as having a single multiplier, a sequence of 9 multiply-and-accumulate (MAC) units, and as a sequence of 81 MAC units, with $|I|=10^5$ random samples per each scenario. These analyses are performed for estimating the accumulated error of a convolution with $3\times3$ and $9\times 9$ filters, respectively. We selected these values because they reflect the size of the convolutional kernels of the DeepCaps \cite{Rajasegaran2019DeepCaps} and CapsNet \cite{Sabour2017dynamic_routing}.

\begin{figure}[h]
	\centering
	\vspace*{-0mm}
	\includegraphics[width=\linewidth]{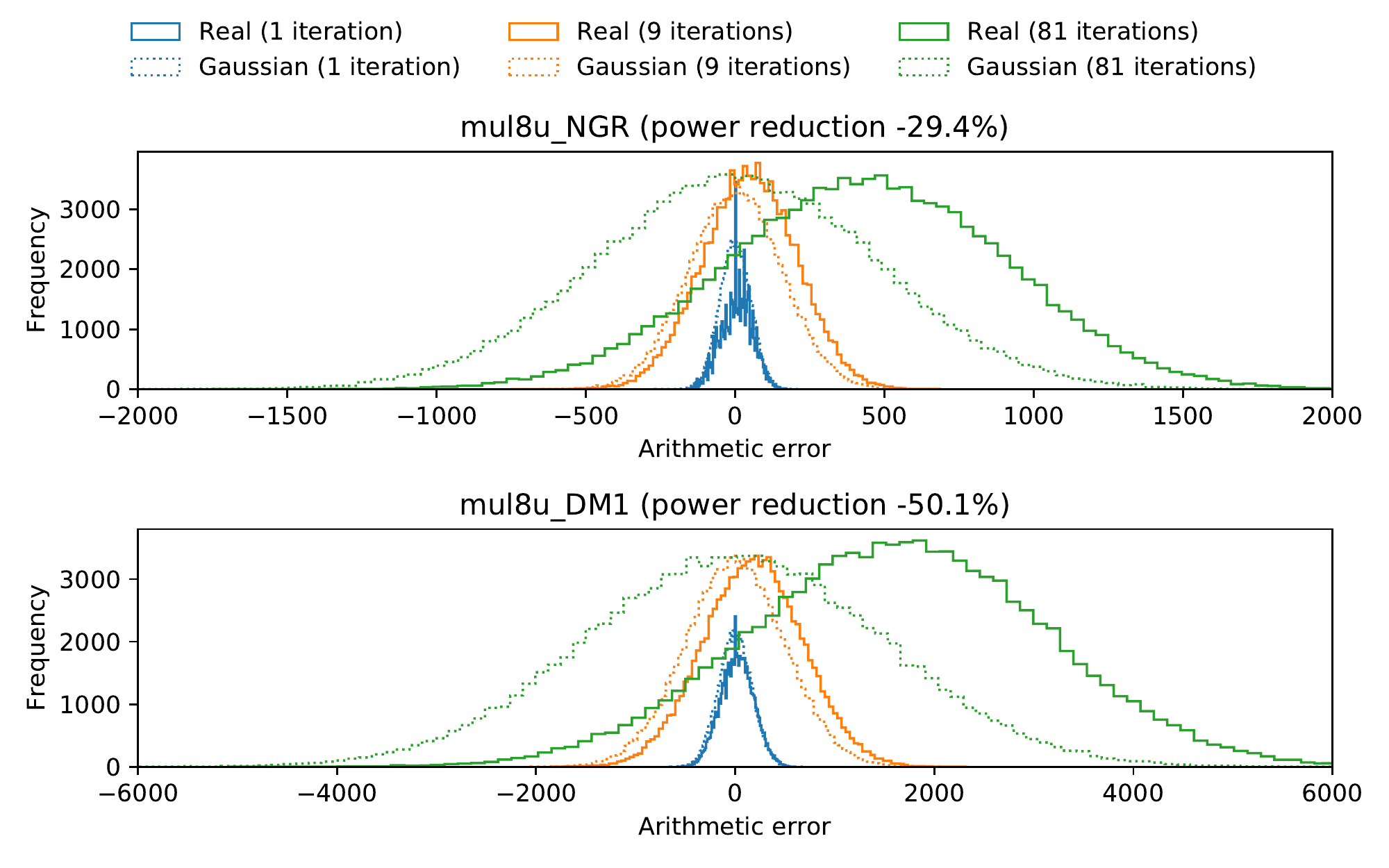}
	\vspace{-5mm}
	\caption{Artimetic error (w.r.t. accurate 8-bit multiplier) distributions and their interpolations by Gaussian noise by having an approximate multiplier from the EvoApprox8B library~\cite{Mrazek2017Lib}. (\textbf{top}) Distribution for the NGR multiplier. (\textbf{bottom}) Distribution for the DM1 multiplier. %\vmR{Vojta there are only 2 multipliers, NGR and DM1. So, instead of ``different types'', say something specific of these two multipliers and why you have selected these two here. Do the others have a similar behavior?}.
	}
	\label{fig:multdist}
	\vspace*{-2mm}
\end{figure}

The majority of the components (31 of 35) has a Gaussian-like distribution of the arithmetic error $\Delta$, with a mean value \textit{m} and a standard deviation \textit{std}.
%Since the mean error is sufficiently close to 0, the error distribution of employing approximate multipliers can be defined by specifying only the \textit{standard deviation} of a Gaussian distribution.
The error distributions of two approximated multipliers\footnote{Since the remaining 29 elements from the EvoApprox8B library~\cite{Mrazek2017Lib} which have a Gaussian-like distribution show a similar behavior, we only report these two examples of approximate multipliers.} from \cite{Mrazek2017Lib} are shown in Fig.~\ref{fig:multdist}.

%The distributions of the arithmetic errors are calculated as having input set $I = (0 \dots 256) \times (0 \dots 255)$. The results reported in the Fig.~\ref{fig:multdist} show that the majority of the components (\vmR{TODO out of TODO}) \vm{the total should be consistent with the previous paragraph} has a Gaussian-like distribution of the arithmetic error $\Delta$. Since the mean error is sufficiently close to 0, the error distribution of employing approximate multipliers can be defined by specifying only the \textit{standard deviation} of a Gaussian distribution.

%In the first analysis of approximate components we focus on approximate multipliers, because the multiplication is widely employed in the computational path. We select 36 components fro EvoApprox8B library~\cite{Mrazek2017Lib} and we analyze the distributions of output error of the approximate multiplier $M'$ compared to accurate 8-bit multiplier $M$~(Fig.~\ref{fig:multdist}), formally
%$$\Delta_{M'} = \{\forall a,b: M'(a,b) - M(a,b)\}. $$
%The distributions are calculated for all inputs of the multiplier (in one iteration) and also accumulated error for 9 and 25 iterations to simulate a convolution by $3\times3$ and $5\times 5$ filters with $10^5$ random samples.  We investigate that the majority of the components (31 of 36) has a Gaussian-like  distribution of $\Delta$. Since the mean error was close to 0, the error distribution can be determined by a \textit{standard deviation} property only.

Modeling a Gaussian noise $\Delta$, when employing $b$-bit fixed-point approximate components in a CapsNet which has floating-point operations, is an open research problem. We propose to adjust the noise w.r.t. the range $R$ of values of a given array $X$. Hence, we introduce the noise magnitude ($NM$) to indicate the standard deviation ($std$) of the noise $\Delta$ scaled w.r.t. $R(X)$, and the noise average ($NA$) to indicate the mean value ($m$) of the noise $\Delta$ scaled w.r.t. $R(X)$.

%The open problem is how to model the Gaussian noise $\Delta$ of fixed-point $b$-bit components in the NN which uses floating-point operations. We propose to stretch the noise to the range of the input tensor $T$ where the noise should be injected. We propose the \textit{noise magnitude (NM)} property that describes standard deviation of relatively to the input range R
\vspace*{-2mm}
$$NM(\Delta_M)=\frac{stdev(\Delta_X)}{R(X)},\ NA(\Delta_M)=\frac{m(\Delta_X)}{R(X)}$$

Since the inputs of components ($I$) employed in CapsNets have typically some specific distribution patterns, the $NM$ of the approximate component is dependent on the application. This implies that the $NM$ can change significantly for different CapsNet models and different dataset used. Hence, we show several experiments for different benchmarks in Sec. \ref{sec:results}. %\vmR{Vojta, I think we should add here which is the consequence that NM is dependent on the application, if there are some limitations, restrictions in our work and/or limitations. Please check if the text in blue in this paragraph is correct}

\vspace*{-1mm}

\subsection{Noise Injection Modeling}
\label{subsec:noise_injection_model}

\vspace*{-.5mm}

%\vmR{TODO: change removing the EP}

%Without any loss of generality, we can model the error source coming from approximate components and bit flips as a Gaussian random noise added to the array $X$ in which we want to inject the error, with a certain probability. 
Based on the above analysis, without loss of generality, we can model the error source coming from approximate components as a Gaussian random noise added to the array $X$ under consideration. 
%The parameters of our noise injection model are:
%\begin{itemize}[leftmargin=*]
    %\item \textbf{Noise Magnitude (NM):} the standard deviation of Gaussian noise normalized to the output range
    %\item \textbf{Error Probability (EP):} the probability of the element $X_i$ being noisy
%\end{itemize}

An error with certain values of $NM$ and $NA$, associated to a given tensor $X$ (i.e., a multidimensional output of a CapsNet operation) with shape $s$ is modelled as in Equation \ref{eq:DeltaX}. The noisy output is denoted as $X'$ in Equation \ref{eq:Xprime}.
\vspace*{-0mm}
\begin{equation}
%\resizebox{.91\linewidth}{!}{%
%    $\Delta_X = \begin{cases}
%        Gauss(s,NM\cdot \mathrm{range}(X)) & \mathrm{if~} Unif(s) < EP \\
%        0 & \mathrm{otherwise}
%    \end{cases}$
%    }
        \Delta_X =  Gauss(s,(NM\cdot R(X))) + (NA \cdot R(X))
    \label{eq:DeltaX}
\end{equation}
\vspace*{-8mm}

\begin{equation}
    X' = X + \Delta_X
    \label{eq:Xprime}
    \vspace*{-0mm}
\end{equation}

Here, $Gauss(s,std)+m$ is a function which generates a tensor of random numbers with shape $s$, following a Gaussian distribution with mean $m$ and standard deviation $std$.  %, and $Unif(s)$ is a function which generates a tensor of random numbers with shape $s$, following a uniform distribution within $[0,1]$.

%For example, a stuck-at fault at the two LSBs of the first weight of a convolutional filter can be modeled by setting $(EP=100\%, NM=0.006)$. Refer to the Fig.~\ref{fig:random_memory} for the $NM$ settings.

\section{ReD-CaNe: Our Methodology for Error Resilience Analysis and Design of Approximate CapsNets}
\label{sec:ERCaNe_methodology}

\begin{figure}[h]
	\centering
	\vspace*{-1mm}
	\includegraphics[width=.95\linewidth]{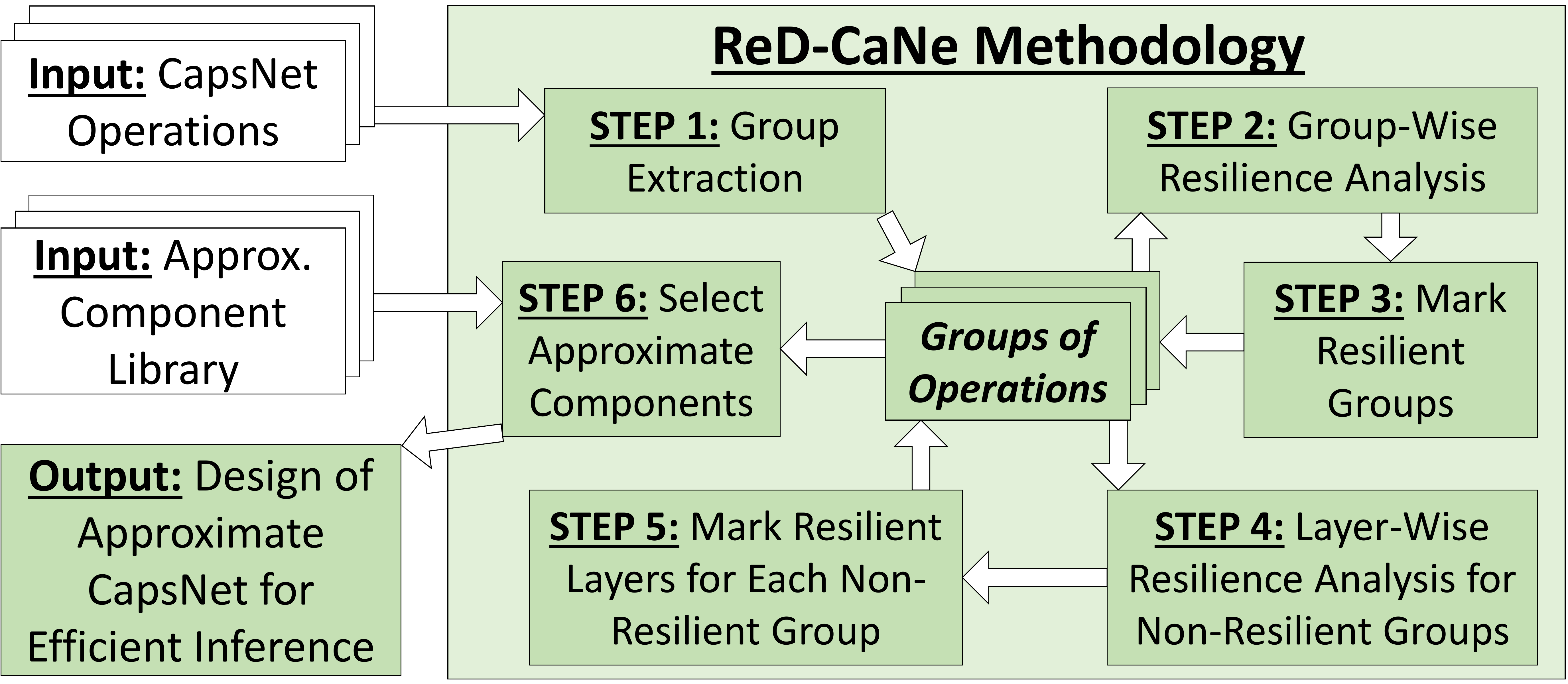}
	\vspace*{-1mm}
	\caption{ReDCaNe: our methodology for resilience analysis and design of Approximate CapsNets.}
	\label{fig:ERCaNe_methodology}
	\vspace*{-1mm}
\end{figure}

Our methodology is composed of 6 steps, as shown in Fig. \ref{fig:ERCaNe_methodology}. Once we identify the lists of arrays in which we want to inject noise, called \textit{Groups}, we apply the noise injection, as described in Sec. \ref{subsec:noise_injection_model}. By monitoring the impact on the test accuracy of different arrays of operations, we can identify the most and the least critical operations in a given CapsNet from the accuracy point of view. Therefore, our ReD-CANE methodology can provide useful guidelines for designing energy-efficient inference, showing the potential to apply approximations to specific layers and operations (i.e., \textit{the more resilient ones}) without significantly sacrificing the accuracy. A step-by-step flow of our methodology is described in the following:

\begin{enumerate}[leftmargin=*]
    \item \textbf{Group Extraction:} We divide the operations of the CapsNet inference into groups, based on the type of operation (e.g., MAC, activation function, softmax or logits update). This step generates the \textit{Groups}.
    \item \textbf{Group-Wise Resilience Analysis:} We monitor the test accuracy drop by injecting noise to different groups.
    \item \textbf{Mark Resilient Groups:} Based on the results of the analysis performed at the Step 2, we mark the more resilient groups. After this step, there are two categories of Groups, the \textit{Resilient} and \textit{Non-Resilient} ones.
    \item \textbf{Layer-Wise Resilience Analysis for Non-Resilient Groups:} For each non-resilient group\footnote{Compared to a layer-wise analysis for each group, by performing such analysis to the non-resilient groups \textit{only}, a considerable amount of unuseful testing can be skipped, and a significant exploration time is saved.}, we monitor the test accuracy drop by injecting noise at each layer.
    \item \textbf{Mark Resilient Layers for Each Non-Resilient Group:} Based on the results of the analysis performed at the Step 4, we mark the more resilient layers.
    \item \textbf{Select Approximate Components:} For each operation, we select approximate components from a given library, based on the resilience measured as the noise magnitude $(NM)$.
\end{enumerate}

Note, a step of resilience analysis consists of setting the input parameters of the noise injection, i.e., $NM$ and $NA$, to add the noise to the selected CapsNet operations, and monitoring the accuracy for the noisy CapsNet.

The output of our methodology is the approximated version of a given CapsNet, which is ready to be executed in a specialized hardware accelerator for inference with approximate components. For the purpose of saving area and energy, we select, for each operation, the approximate components, from a given library, that correspond to their level of resilience. Hence, more aggressive approximations are selected for more resilient operations, without significantly affecting the classification accuracy of the CapsNet inference.

%Indeed, based on the level of resilience of the groups and the layers, we select more approximate components (for the purpose of saving area, energy), from a given library, to compute the operations marked as more resilient, without significantly affecting the accuracy at the inference stage.% Examples of approximations are quantization and the usage of approximate components (e.g., adders, multipliers) in specialized hardware accelerators.

\vspace*{-0mm}
\section{Experimental setup}
\label{sec:setup}
\vspace*{-0mm}

\begin{figure}[h]
	\centering
	\vspace*{-1mm}
	\includegraphics[width=.8\linewidth]{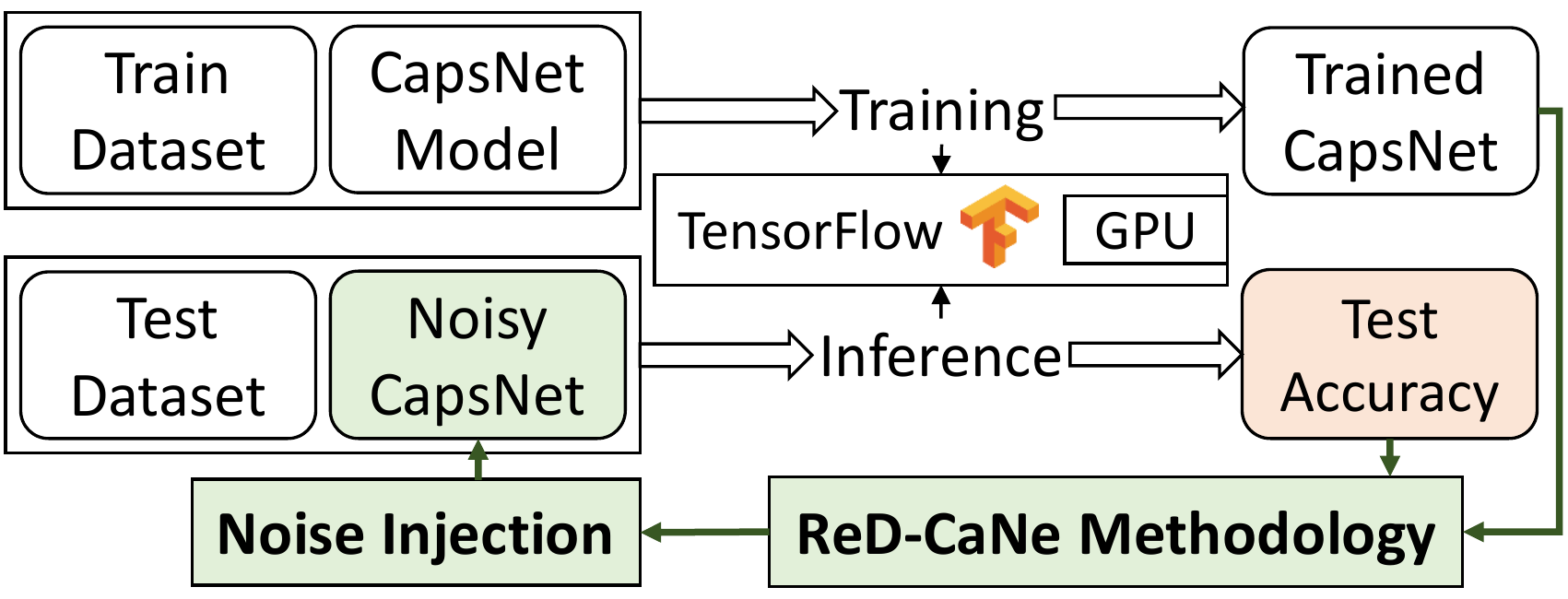}
	\vspace*{-1mm}
	\caption{Experimental setup for our simulations.}% \vm{add more details based on Vojta's figure, but maybe they can just be discussed in the text}}
	\label{fig:exp_setup}
	\vspace*{-1mm}
\end{figure}

The experimental setup is shown in Fig. \ref{fig:exp_setup}. We train a given CapsNet model for a given dataset using TensorFlow \cite{Google2016TensorFlow}, running on two Nvidia GTX 1080 Ti GPUs. The trained model serves as an input to our ReD-Cane methodology. The noise is injected to the arrays and then the accuracy is monitored to identify the resilience of the operations.

%\vm{When we descibe the CapsNet models, we can say that making the tests for DeepCaps is more comprehensive, but we keep on having results also for the CapsNet (Sabour) because it is widely known and adopted by the community, as it is the first and original paper about CapsNets.}

\subsection{CapsNet Models and Datasets}
\vspace*{-.5mm}

We test our methodology on two networks, the DeepCaps \cite{Rajasegaran2019DeepCaps} and the original CapsNet \cite{Sabour2017dynamic_routing}. We use the datasets CIFAR-10 \cite{Krizhevsky2009CIFAR}, SVHN \cite{Netzer2011SVHN}, MNIST \cite{LeCun1998MNIST}, and Fashion-MINST \cite{Xiao2017Fashion-MNIST}, to classify generic images, house numbers, handwritten digits, and fashion clothes, respectively. The accuracy results obtained by training these networks for different datasets are reported in Table \ref{tab:accuracy_accurate}. Table \ref{tab:groups} shows the partition of the CapsNet operation into groups, which is then used for the group extraction step.

%\vm{Schematic figure, more detailed than the one in the introduction but similar. We inserted a specific node in a tensor, where it's possible to set the noise. Then, having defined the points in the computational graph where we can add the noise, our framework can add the noise. At the output, we measure the accuracy. Then, a table or figure which describes the division of the operation into groups, having said that we described in detailed the architecture of the DeepCaps before (either earlier in this section or in another section). Explain implementation issues like protobuf etc.}

\begin{table}[h]
\vspace*{-1mm}
\captionsetup{font=small}
\caption{Classification accuracy results using accurate multipliers.% \vmR{TODO: correct the values of accuracy 92.74 97.56 99.72 92.88 99.67. Actual values are 90.46 96.62 99.22 89.58 99.28, fixable after the submission}
}
\label{tab:accuracy_accurate}
\vspace*{-1.5mm}
\centering
\resizebox{0.6\linewidth}{!}{%
\centering\renewcommand{\arraystretch}{0.9} 
%\resizebox{0.6\linewidth}{}{}
\begin{tabular}{c c c}
\toprule
\textbf{Architecture} & \textbf{Dataset} & \textbf{Accuracy} \\ \midrule
\multirow{3}{*}{DeepCaps \cite{Rajasegaran2019DeepCaps}} & CIFAR-10 & 92.74 \\
                                    & SVHN & 97.56 \\
                                    & MNIST & 99.72 \\\midrule
\multirow{2}{*}{CapsNet \cite{Sabour2017dynamic_routing}} & Fashion-MNIST & 92.88 \\
                                    & MNIST & 99.67\\\bottomrule
\end{tabular}%
}
\vspace*{-2mm}
\end{table}

\begin{table}[h t]
\captionsetup{font=small}
\caption{Grouping the operations of the CapsNet inference.}
\label{tab:groups}
\vspace*{-1.5mm}
\resizebox{\linewidth}{!}{%
\begin{tabular}{|c|l|l|}
\hline
\textbf{\#} & \textbf{Group Name} & \textbf{Description} \\ \hline
%1 & Input & Pixels of the input image \\ \hline
%2 & Weights & Convolutional kernels \& weights for element-wise mult. \\ \hline
1 & MAC Outputs & Outpus of the matrix multiplications \\ \hline
2 & Activations & Output of the activation functions (RELU or SQUASH) \\ \hline
3 & Softmax & Results of the softmax (\textit{k} coefficnents in dynamic routing) \\ \hline
4 & Logits Update & Update of the logits (\textit{b} coefficients in dynamic routing) \\ \hline
\end{tabular}%
}
\vspace*{-2mm}
\end{table}

% \begin{table}[ht]
%     \centering
%     \caption{Computational complexity of the groups in the computational path in the DeepCaps architecture}
%     \label{tab:my_label}
%     \resizebox{\linewidth}{!}{
%     \begin{tabular}{c c c c c c c c}\toprule
%         \bf \# & \bf Group & \# Mults & \# Adds & \# Divs & \# Exps & \# Sqrts & Rel. energy \\\midrule
% %1 & Input &  \\
% %2 & Weights &  \\
% 1 & MAC Outputs & \\
% 2 & Activations &  \\
% 3 & Softmax &  \\ 
% 4 & Logits Update &  \\ \bottomrule
%     \end{tabular}}
% \end{table}

\subsection{TensorFlow Implementation}

\vspace*{-.5mm}

The proposed methodology is implemented in TensorFlow \cite{Google2016TensorFlow}. First, the network is trained using standard approaches. We modified the computational graph in \textit{protobuf} format by including our noise injection model in the \textit{Graph tool}. We implemented a specialized node for the noise injection, where the values of $NA$ and $NM$ can be specified as inputs to this node. Hence, for each node $\tau$, a new set of nodes $T$ is added to the graph. The nodes in $T$ have the same shape as $\tau$ and they consist of the set of operations for adding a Gaussian noise with $std = NM \cdot R(\tau)$ and $m = NA \cdot R(\tau)$, given the range $R$ of the node $\tau$.

%For each node $T$ in this list new nodes (operations) are added: the range $R(T)$ of the tensors is calculated, a Gaussian noise with the same shape as $T$ is generated and multiplied by $NM\cdot R(T)$ values and $NA \cdot R(T)$ is added.

%The accuracy of the updated graphs is evaluated and analyzed using proposed ReD-CaNe methodology.

%\vmR{Explain implementation issues like protobuf etc.}

%\vmR{TODO: Vojta describe the implementation. PRoblem of IFs in TF }

%Without any loss of generality, we can model the error source coming from approximate components and bit flips as a Gaussian random noise added to the array $X$ in which we want to inject the error, with a certain probability. The noise injection model is devised in Figure \ref{fig:inject_noise}. It has three inputs, the input array $X$ and two other parameters which can be arbitrarily set based on which type of error we want to model. In this way, we are able to simulate the effect of several different sources of error.

%\vm{TODO: Discuss the quantization and that the error (Stdev) must be specified with respect to the output range - Noise magnitude property}

%The intermediate results $Z$ and $K$, as well as the noisy output $Y$, are generated as described in Eqs. \ref{eq:noise_Z}, \ref{eq:noise_K}, \ref{eq:noise_Y}, respectively.

%\vm{TODO: Correct the equations.}% the error is: Gaussnoise(stddev=NM*input range) if random() < ep; 0 otherwise

% Description of the GPU and the tools like TensorFlow, and description of the CapsNets models and datasets used, and the block to add the noise.

\vspace*{-1mm}

\subsection{Approximate Multiplier Library}

\vspace*{-.5mm}

We use the EvoApprox8b library, which consists in 35 8-bit unsigned components. We select 8-bit wordlength since it was shown to be enough accurate in the computational path of CapsNets~\cite{Marchisio2019CapsAcc}. 

\vspace*{-1mm}

\section{Experimental Results}
\label{sec:results}

\vspace*{-1mm}

\subsection{Detailed Analysis for the CIFAR-10 Dataset}

\vspace*{-.5mm}

As a case study analysis, we report detailed results for the DeepCaps on the CIFAR-10 datasets. The results for other benchmarks are reported in Section~\ref{sec:allbench}.

\begin{figure}[h]
	\centering
	\vspace{-1mm}
	\includegraphics[width=\linewidth]{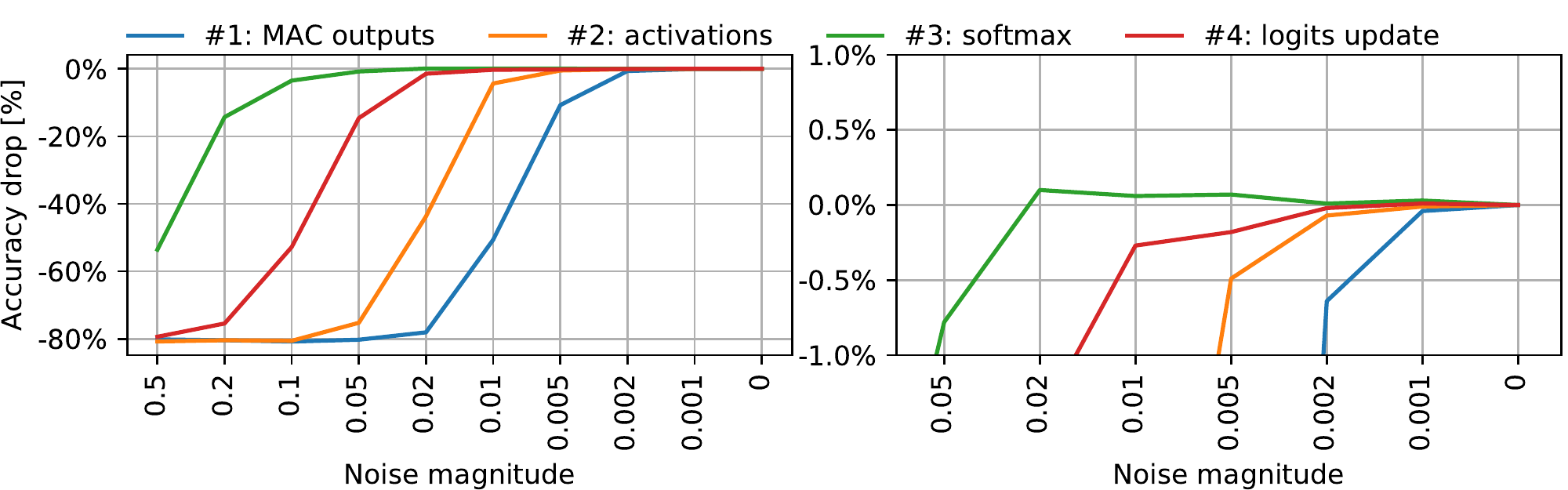}
	\vspace{-5mm}
	\caption{Group-wise resilience for the CIFAR-10 dataset. (\textbf{left}) Complete results. (\textbf{right}) Zoomed view of the accuracy drop, centered at $0\%$.}
	\label{fig:accurate_c10}
	\vspace{-1mm}
\end{figure}

For the following analyses, we used a $NM \in [0.5 \dots 0.001]$. To analyze the general case of error resilience, we selected the average error $NA=0$. 
%\vmR{specify somewhere that we have used mean=0 in the experiments, justified by the fact that we are using an approximated model of the distribution of the approximate multiplier error.}
In the experiment for the \textbf{Step 2} of our methodology, we inject the same noise to every operation within a group, while keeping the other groups accurate. From the results shown in Fig.~\ref{fig:accurate_c10}, we notice that the \textit{Softmax} and the \textit{Logits update} groups are more resilient than \textit{MAC outputs} and \textit{Activations}, because the CapsNet accuracy starts to decrease with a correspondent lower $NM$. Note, for low $NM$, the noise injection slightly increases the accuracy due to regularization, with a similar effect as the dropout \cite{Srivastava2014Dropout}.% effect \vmR{cite}.

\begin{figure}[h]
	\centering
	\vspace{-0mm}
	\includegraphics[width=\linewidth]{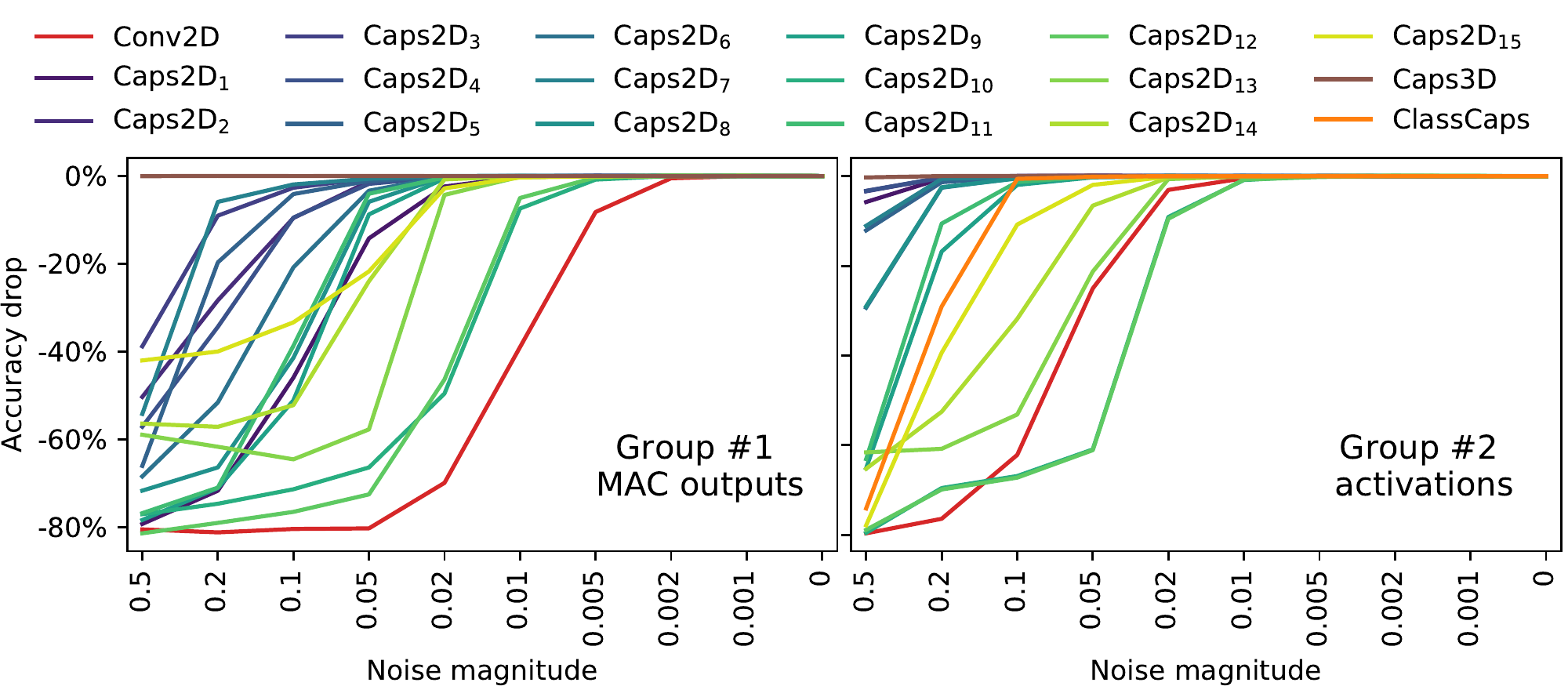}
	\vspace{-5mm}
	\caption{Layer-wise analysis of the non-resilient groups for CIFAR-10 dataset. The noise is injected in (\textbf{left}) the MAC outputs and (\textbf{right}) the activations.}
	\label{fig:layerwise_c10}
	\vspace{-3mm}
\end{figure}

%\subsection{Layer-Wise Resinience of Non-Resilient Groups}
In Fig.~\ref{fig:layerwise_c10}, we analyze the resilience of each layer of the non-resilient groups (i.e., \textit{MAC outputs} and \textit{Activations}).
We notice that the first convolutional layer is the least resilient, followed by the layers in the middle. Moreover, the Caps3D layer is the most resilient one. Since this layer is the only convolutional layer that employs the dynamic routing algorithm, we correlate the higher resilience to the iterations performed in this layer, because \textit{the coefficients are updated dynamically at run-time, thus they can adapt to the noise}.

%\subsection{Resilience of weights}
%\begin{figure}[H]
%	\centering
%	\includegraphics[width=\linewidth]{img/memory_c10.pdf}
%	\caption{Approximation of weights for CIFAR-10.}
%	\label{fig:memory_c10}
%\end{figure}
%
%The last experiment analyses what would happened whether one would combine error in both least error resilient groups (Fig.~\ref{fig:2dzoom_c10}). We can see that the error can be injected into both groups with a small impact on the overall accuracy drop while the activation layers are less sensitive for the errors. 

%\begin{figure}[H]
	%\centering
	%\includegraphics[width=\linewidth]{img/2dzoom_c10.pdf}
	%\caption{Accuracy drop for CIFAR-10 while a different noise was injected to MAC outputs and the Activations. (\textbf{left}) Completed result. (\textbf{right}) Configurations having accuracy drop $\leq -5\%$.}
	%\label{fig:2dzoom_c10}
%\end{figure}

\vspace*{-1.5mm}
\subsection{Evaluating the Selection of Approximate Components}
\vspace*{-.5mm}

\begin{figure}[h]
    \centering
    \vspace*{-1mm}
    \includegraphics[width=\linewidth]{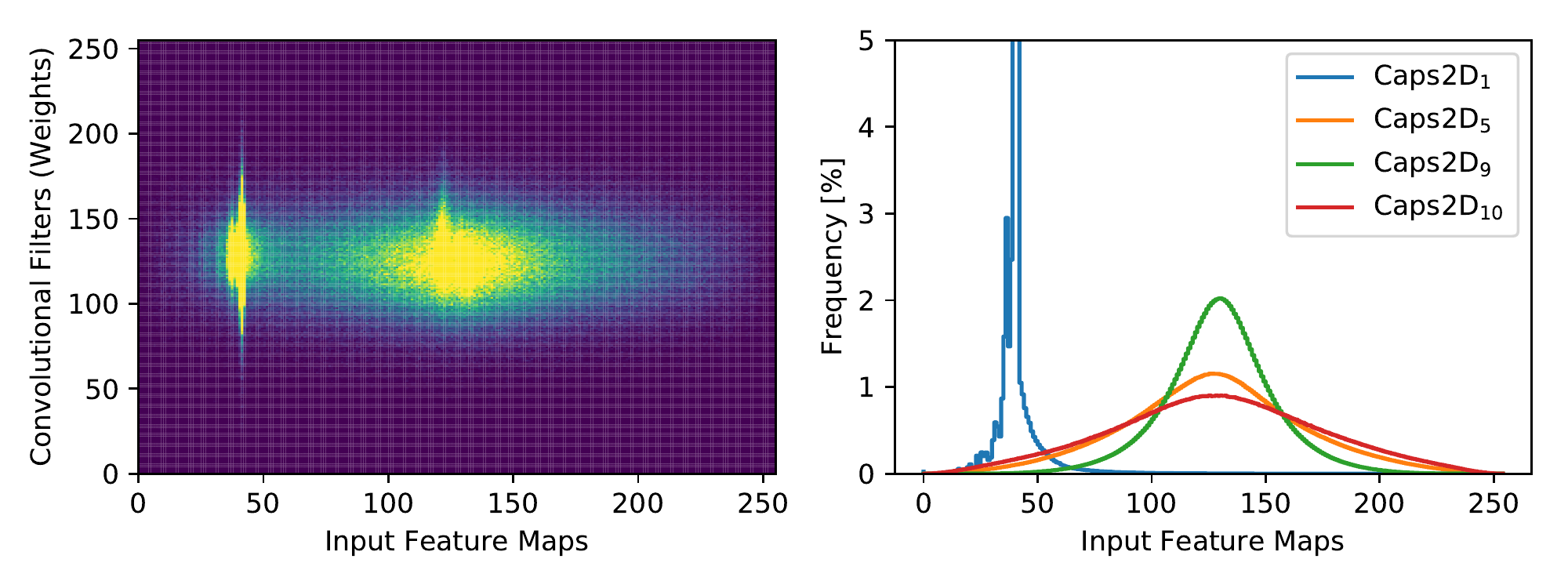}
    \vspace*{-5mm}
    \caption{(\textbf{left}) Distribution of $10^6$ random samples from the inputs of the convolutions in the DeepCaps for the CIFAR-10 dataset. (\textbf{right}) A focus on some layers, showing the peak in the first Caps2D layer. %\vmR{Vojta please use the same convention of names of the layers as in Fig 10, with the pedix as index. For example, Caps2D$_1$}
    }
    \label{fig:dist}
    \vspace*{-1mm}
\end{figure}

The choice of the approximate 
component for each operation depends on the level of $NM$ corresponding to a tolerable accuracy loss, which is typically null or very low. Recalling Eq.~\ref{eq:arithmetic_error}, the parameters $NM$ and $NA$ are dataset dependent because their values change accordingly to the input range $R$. In our case study (DeepCaps for CIFAR-10), we select a subset of $10^6$ elements from the inputs of every \textit{Conv2D} layers of the DeepCaps, with its corresponding distribution (frequency of occurrence) shown in Fig.~\ref{fig:dist} (left). The distribution is approximately Gaussian, but there is a peak between $40$ and $50$ for the input feature maps, which is caused by a specific distribution of the input dataset. Indeed, the peak occurs in the first Caps2D layer, as shown in Fig.~\ref{fig:dist} (right).

%The $NM$ and $NA$ parameters depend on a selected subset of input data $I$ (see Eq.~\ref{eq:arithmetic_error}. For the correct evaluation,  we analyzed the behavior of the input data for convolutions of DeepCaps network performing the inference of images in the CIFAR-10 dataset  in the first experiment. 

%We select $10^6$ samples of the inputs of all \textit{Conv2D} operations and the distribution (frequency of occurence) is shown in Fig.~\ref{fig:dist}. The distribution of the inputs and filters is almost Gaussian. However, there is a peak of inputs between $40-50$ value. This peak is caused by distribution of input images and it occurs in the first and in the second layer only.

Hence, we measure the $NM$ and $NA$ parameters of the selected multipliers in the library (Tab.~\ref{tab:mults}). We use two different input distributions, the \textit{modeled} one that is based on random inputs generated with a uniform distribution, and the \textit{real} one, which is based on the input distribution previously shown in Fig.~\ref{fig:dist}. Note, these values slightly differ, because the $NM$ and $NA$ parameters are dataset dependent. The major differences are due to an overestimation of the $NM$ and $NA$ by our modeled distribution. Therefore, the selection of approximate components based on our models can be systematically employed for designing approximate CapsNets.

%--- Generic that uses random inputs generated with uniform distribution and Measured based on the input dataset $I$ shown in Fig.~\ref{fig:dist}. As we can see, these values differs for the selected datasets. Therefore, the $NM$ and $NA$ parameters are dataset-dependend. 

\begin{table}[h]
\vspace*{-1mm}
    \caption{Power, area, and noise magnitude, computed with a modeled input dataset (with uniform distribution) and with a real input distribution $I$, for different{\footnotesize $^*$} approximated multipliers from the EvoApprox8B library \cite{Mrazek2017Lib}. }
    \label{tab:mults}
    \vspace{-1mm}
    \centering
    \resizebox{\linewidth}{!}{
    \renewcommand{\arraystretch}{1.0} 
    \begin{tabular}{c|cc|cc|cc}\toprule
    \bf Multiplier & \bf Power & \bf Area & \multicolumn{2}{c|}{\bf Modeled $\Delta_X$} & \multicolumn{2}{c}{\bf  Real $\Delta_X$} \\
    mul8u\_ & $\mu$W & $\mu$m$^2$ & $NA$ & $NM$  & $NA$ & $NM$ \\\midrule
1JFF & 391 (-0\%) & 710 (-0\%) & 0.0000 & 0.0000 & 0.0000 & 0.0000\\\midrule
14VP & 364 (-7\%) & 654 (-8\%) & 0.0000 & 0.0001 & 0.0000 & 0.0001\\
GS2 & 356 (-9\%) & 633 (-11\%) & 0.0004 & 0.0017 & 0.0001 & 0.0013\\
CK5 & 345 (-12\%) & 604 (-15\%) & 0.0000 & 0.0002 & 0.0000 & 0.0002\\
7C1 & 329 (-16\%) & 607 (-14\%) & 0.0011 & 0.0033 & 0.0007 & 0.0026\\
96D & 309 (-21\%) & 605 (-15\%) & 0.0035 & 0.0077 & 0.0020 & 0.0051\\
2HH & 302 (-23\%) & 542 (-24\%) & -0.0001 & 0.0007 & -0.0001 & 0.0007\\
NGR & 276 (-29\%) & 512 (-28\%) & 0.0001 & 0.0008 & 0.0002 & 0.0009\\
19DB & 206 (-47\%) & 396 (-44\%) & 0.0010 & 0.0019 & 0.0010 & 0.0021\\
DM1 & 195 (-50\%) & 402 (-43\%) & 0.0003 & 0.0025 & 0.0005 & 0.0025\\
12N4 & 142 (-64\%) & 390 (-45\%) & 0.0018 & 0.0054 & 0.0019 & 0.0056\\
1AGV & 95 (-76\%) & 228 (-68\%) & 0.0027 & 0.0080 & 0.0026 & 0.0117\\
YX7 & 61 (-84\%) & 221 (-69\%) & 0.0484 & 0.0741 & 0.0268 & 0.0347\\
JV3 & 34 (-91\%) & 111 (-84\%) & 0.0021 & 0.0267 & -0.0028 & 0.0301\\
QKX & 29 (-93\%) & 112 (-84\%) & 0.0509 & 0.0736 & 0.0293 & 0.0350\\
\bottomrule
\multicolumn{7}{l}{$^*$We have randomly selected 14 components, representative for the complete library.}
    \end{tabular}
    }
    \vspace{-5mm}
\end{table}

%\subsection{Approximate Component Selection}
%\vmR{Vojta, we should describe the step 6 of the methodology. Based on the NM that we have when the accuracy drops significantly for a certain group or layer, we select the correspondent approx multiplier. Give an example in a table for the selection.}

\subsection{Testing our Methodology on Different Benchmarks}\label{sec:allbench}
\vspace{-.5mm}

\begin{figure}[h]
\vspace{-1mm}
	\centering
	\includegraphics[width=\linewidth]{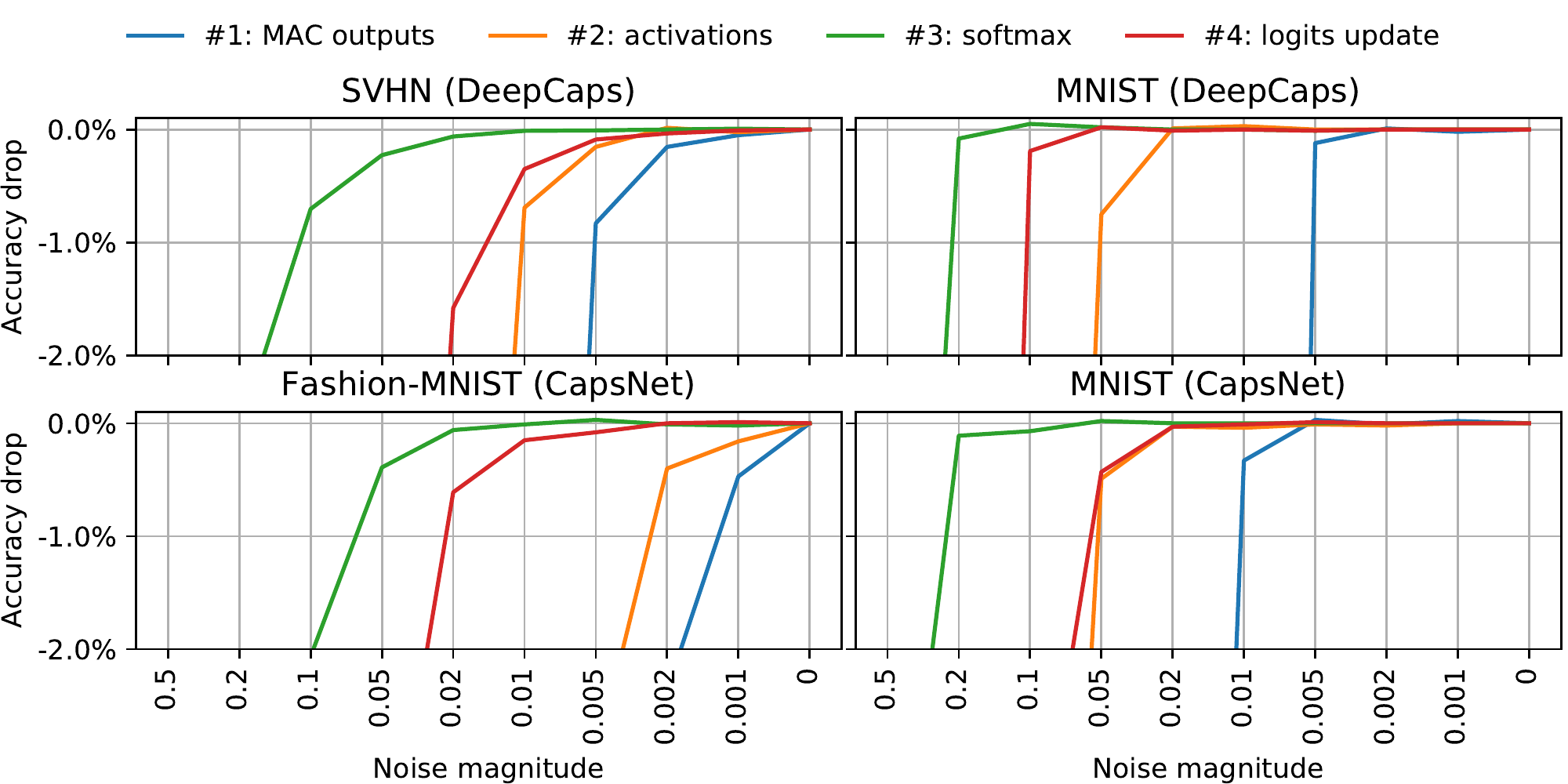}
	\vspace{-5mm}
	\caption{Group-wise resilience for different benchmarks.}% \vmR{zoomed to 0 --- -2\% range}.}
	\label{fig:accurate_all}
	\vspace{-3mm}
\end{figure}

We apply our methodology to the other benchmarks. The results coming from the resilience analysis of the Step 2 are shown in Fig.~\ref{fig:accurate_all}. A key property that we can observe is that MAC outputs and activations are less resilient than the other two groups. Moreover, we noticed that the logits update on the CapsNet \cite{Sabour2017dynamic_routing} for MNIST (bottom right) is slightly less resilient than the same group on the DeepCaps \cite{Rajasegaran2019DeepCaps} for MNIST (top right), because the CapsNet has only one layer that performs Dynamic routing, while the DeepCaps has two. %Add something based on the new figure

\subsection{Results Discussion}
\vspace{-.5mm}

From our analyses, we can derive that the CapsNets have interesting resilience properties. A key observation, valid for every benchmark, is that the layers computing the the dynamic routing (ClassCaps and Caps3D), and the corresponding groups of operations (softmax and logits update) are more resilient than others. Such outcome is attributed to a common feature of the dynamic routing. The values of the involved coefficients (logits $b$ and coupling coefficients $k$, see Fig.~\ref{fig:routing_by_agreement}) are updated dynamically, thereby adapting to the injected noise. Hence, more aggressive approximations can be tolerated for these computations.

\vspace{-2mm}
\section{Conclusion}
\vspace{-1mm}

We proposed a systematic methodology for analyzing the resilience of CapsNets under approximation errors that can provide foundation to design approximate CapsNet hardware. We designed an error injection model, which accounts for the approximation errors. We modeled the errors of applying approximate multipliers in the computational units of CapsNet accelerators. We systematically analyzed the (group-wise and layer-wise) resilience of the operations and designed approximated CapsNets, based on different resilience levels. We showed that the operations in the dynamic routing are more resilient to approximation errors. Hence, more aggressive approximations can be adopted for these computations, without sacrificing the classification accuracy much. Our methodology provides the first step towards real-world approximate CapsNets to realize their energy-efficient inference.

\section*{Acknowledgments}
This work has been partially supported by the Doctoral College Resilient Embedded Systems which is run jointly by TU Wien's Faculty of Informatics and FH-Technikum Wien, and partially supported by the Czech Science Foundation project 19-10137S.

\bibliographystyle{abbrvnat}

\begin{refsize}
\bibliography{main.bib}
\end{refsize}

\end{document}